\documentclass[runningheads]{llncs}
\usepackage{float}
\usepackage{graphicx}
\usepackage{stmaryrd}
\usepackage{xcolor}
\def\boxleft{\textbf{\textrm{box\_left}}}
\def\boxabove{\textbf{\textrm{box\_above}}}
\def\boxbelow{\textbf{\textrm{box\_below}}}

\begin{document}
\title{Unified Line and Paragraph Detection by Graph Convolutional Networks\thanks{Work done during the first author's internship at Google Research.}}
%
%
\author{Shuang Liu\inst{1} \and
Renshen Wang\inst{2} \and
Michalis Raptis\inst{2} \and Yasuhisa Fujii\inst{2}}
\authorrunning{S. Liu et al.}
%
\institute{University of California, San Diego\\
\email{s3liu@eng.ucsd.edu}
\and
Google Research\\
\email{\{rewang,mraptis,yasuhisaf\}@google.com}}
\maketitle              
\begin{abstract}
We formulate the task of detecting lines and paragraphs in a document into a unified two-level clustering problem. Given a set of \emph{text detection boxes} that roughly correspond to words, a text line is a cluster of boxes and a paragraph is a cluster of lines. These clusters form a two-level tree that represents a major part of the layout of a document. We use a graph convolutional network to predict the relations between text detection boxes and then build both levels of clusters from these predictions. Experimentally, we demonstrate that the unified approach can be highly efficient while still achieving state-of-the-art quality for detecting paragraphs in public benchmarks and real-world images.

\end{abstract}

\section{Introduction}

Document layout extraction is a critical component in document analysis systems. It includes pre-OCR layout analysis \cite{tesseract} for finding text lines and post-OCR layout analysis \cite{semantic_structure} \cite{post_ocr} for finding paragraphs or higher level entities. These tasks involve different levels of document entities -- line-level and paragraph-level -- both are important for OCR and its downstream applications, and both have been extensively studied. Yet, to the best of authors' knowledge, there has been no study that tries to find text lines and paragraphs at the same time.

Graph convolutional networks (GCNs) are becoming a prominent type of neural networks due to their capability of handling non-Euclidean data \cite{survey_gnn}. They naturally fit many problems in OCR and document analysis, and have been applied to help form lines \cite{adaptive_boundary} \cite{relatext} \cite{deep_relational}, paragraphs \cite{post_ocr} or other types of document entities \cite{rope}. Besides the quality gain from these GCN models, another benefit from these approaches is that we can potentially combine all the machine learning tasks and build a single, unified, multi-task GCN model.

In this paper, we propose to apply a multi-task GCN model in OCR text detection to find both text lines and paragraphs in the document image. Compared to separated models dedicated to specific tasks, this unified approach has some potential advantages:
\begin{itemize}
\item System performance. Running a single step of model inference is usually faster than running multiple steps.

\item Quality. A multi-stage system often suffers from cascading errors, where an error produced by a stage will cause ill-formed input for the next stage. Pruning stages can reduce the chance for this type of errors.

\item Maintainability. The overall system complexity can be reduced, and it is easier to retrain and update a unified model than multiple models.
\end{itemize}

There are also potential drawbacks on this approach. For example, some post-OCR steps \cite{post_ocr} need to be moved to pre-OCR, which reduces the available input signals to the model.

The rest of this paper is organized as follows. Section 2 reviews the related work. Section 3 presents our proposed method, where the problem formulation and the modeling solution are discussed in details. Section 4 discusses potential drawbacks and limitations of this approach. Section 5 presents experimental results. Finally, section 6 concludes the paper with suggestions for future work.

\section{Related Work}

\subsection{Text Line Detection}

The first step of OCR is text detection, which is to find the text words or lines in the input image. For documents with dense text, lines are usually chosen over words for more reliable performance \cite{rethinking}. Due to the high aspect ratios of text lines, an effective way to find them is to first detect small character-level boxes, and then cluster them into lines using algorithms like Text Flow~\cite{textflow}. 

\begin{figure}
    \centering
    \includegraphics[width=0.8\linewidth]{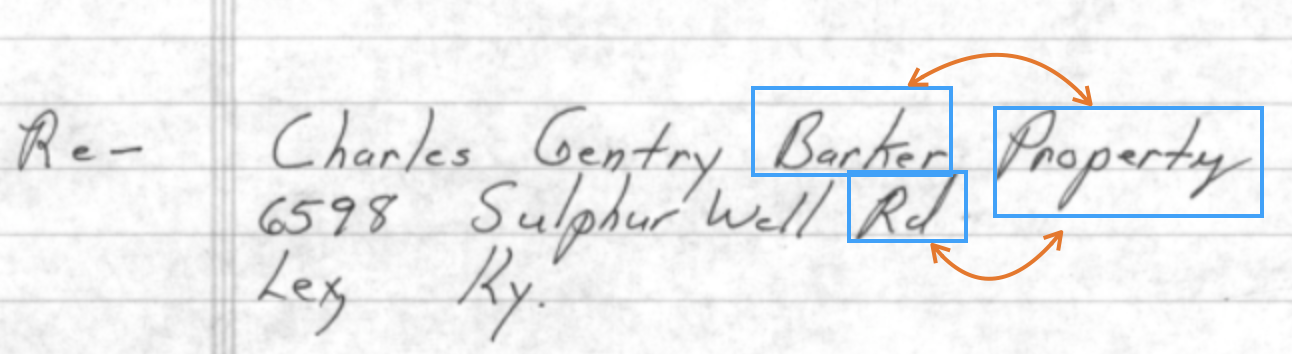}
    \caption{A difficult handwriting example for Text Flow clustering --- the upper positive edge has a larger distance between the pair of words than that of the lower negative edge, which may cause the algorithm to choose the wrong edge.}
    \label{fig:handwriting}
\end{figure}

While Text Flow is effective for most printed documents, handwritten text can be challenging for the network flow model. As shown in Figure~\ref{fig:handwriting}, handwritten words can be sparser and more scattered than printed ones, and may cause difficulty for Text Flow, as well as for connected component based post-processing like \cite{unconstrained}. A sophisticated graph clustering algorithm based on superpixels is proposed in \cite{two_stage} for better performance with historical/handwritten documents, yet with potential limitations on line curvilinearity and word spacings.

We think graph neural network based approaches like \cite{adaptive_boundary} \cite{relatext} \cite{deep_relational} are in a better position to solve the handwritten text detection problem, or more difficult layout problems in general.

\subsection{Paragraph Detection}

PubLayNet \cite{publaynet} provides a large data set containing paragraphs as well as baseline object detection models trained to detect these paragraphs. While the image based detection models can work well on this data set, it is shown in \cite{post_ocr} that they are less adaptable to real-world applications with possible rotations and perspective distortions, and require more training data and computing resources. For example, the detection models in \cite{publaynet} are around 1GB in size, while the GCN models based on OCR outputs in \cite{post_ocr} are under 200KB each.

So once again, graph neural network based approaches have significant advantages in the layout problem. Note that in \cite{post_ocr}, two GCN models take OCR bounding boxes as input to perform operations on lines. If these lines are formed by another GCN model, there will be totally three GCN models in the system, which leaves opportunities for unification and optimization.

\section{Proposed Method}

In this paper, we propose a unified graph convolutional network model to form both text lines and paragraphs from word-level detection results.

\subsection{Pure Bounding Box Input}
\label{sec:gcn_boxes}

Our GCN model takes only geometric features from bounding boxes like \cite{post_ocr}, rather than using RRoI (rotated region of interest) features from input images \cite{deep_relational}. Not only does it greatly reduce the model size and latency, but it also makes the model generalize better across domains --- as verified by the synthetic-to-real-world test in \cite{post_ocr}.

Not coincidentally, a street view classification model proposed in \cite{bounding_boxes} --- which can be viewed as another type of layout analysis --- is discovered to perform better with pure bounding boxes.

We use word-level boxes that are obtained by grouping character-level boxes. Since the difficulty of Text Flow shown in Figure~\ref{fig:handwriting} mainly resides in the connections between words, character-to-word grouping is easier and can be performed with a geometric heuristic algorithm at a high accuracy.
 
\subsection{Problem Statement}

The input of our algorithm will be a set of $n$ rotated rectangular boxes as well as $m$ undirected edges between them. 

\begin{figure}
    \centering
    \includegraphics[height=0.22\linewidth]{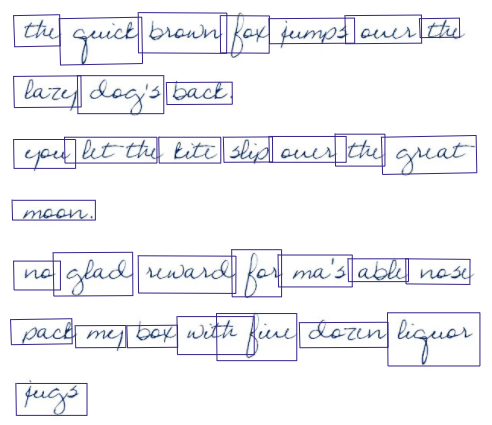}\hspace{2mm}
    \includegraphics[height=0.22\linewidth]{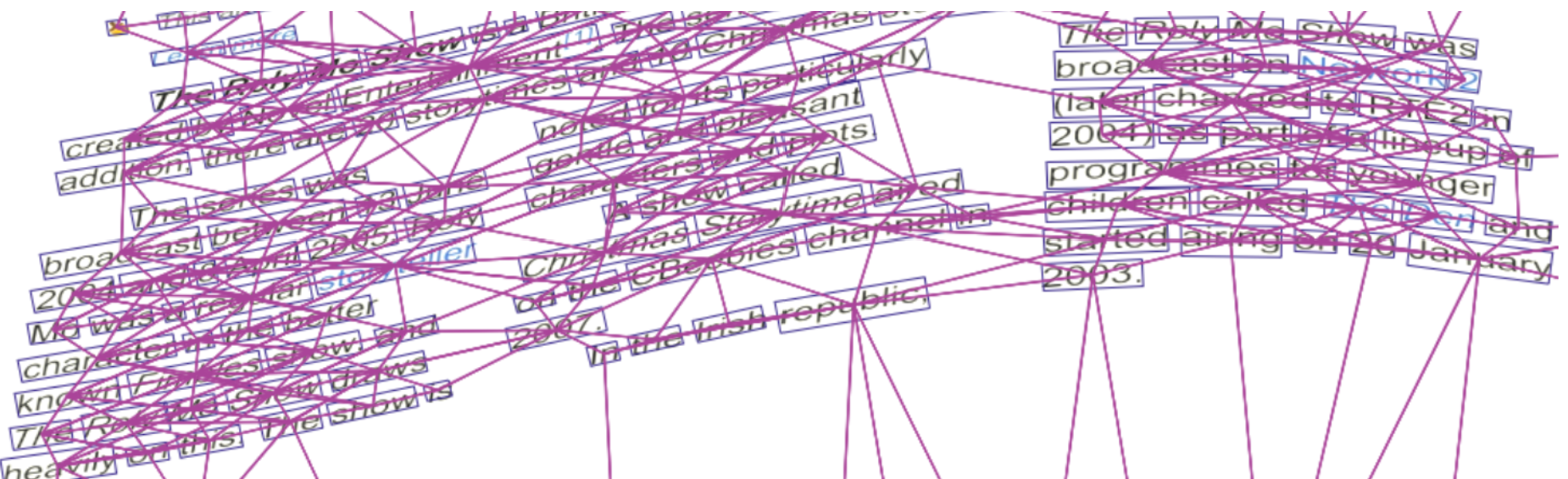}
    \caption{Left: GCN boxes (blue) which do not necessarily correspond to words or characters. Right: GCN edges (purple) over GCN boxes (blue).}
    \label{fig:gcn_boxes_and_edges}
\end{figure}

\begin{itemize}
\item
Each box is represented by five real numbers $(x_i, y_i, w_i, h_i, a_i)$ where $(x_i, y_i)$ is the coordinate of its upper left corner, $w_i$ is its width, $h_i$ is its height, and $a_i$ is its angle. We will call these \emph{GCN boxes}, since they will be used as nodes by a GCN. These boxes correspond to word-level text regions in a document; an example is shown on the left of Figure~\ref{fig:gcn_boxes_and_edges} --- note that each GCN box does not necessarily correspond to a word or a character in the usual sense. While our approach is largely agnostic to the specification of the boxes, the approximate word level boxes discussed in Section~\ref{sec:gcn_boxes} are preferred for their good accuracy and efficiency.

\item
The edges connect pairs of boxes that are close to each other. We will call these edges \emph{GCN edges} since they will be used as edges by a GCN. An example is shown on the right of Figure~\ref{fig:gcn_boxes_and_edges}. We describe how the edges are constructed in Section~\ref{sec:gcn_edges}. 
\end{itemize}

The output of our algorithm should be a two-level clustering of the input: ideally, the first level clusters GCN boxes into lines, and the second level clusters lines into paragraphs.

\subsection{Main Challenge}

For efficiency considerations, the GCN edges should be constructed in a way such that its size grows linearly with the number of GCN boxes --- we will call such a GCN a linear-sized GCN. We discuss a way of constructing a linear-sized graph for a GCN in Section~\ref{sec:gcn_edges}. However, this requirement introduces a problem: as we will explain shortly, paragraph is a global property, while a linear-sized GCN makes only local predictions.

\begin{figure}
    \centering
    \includegraphics[width=0.7\linewidth]{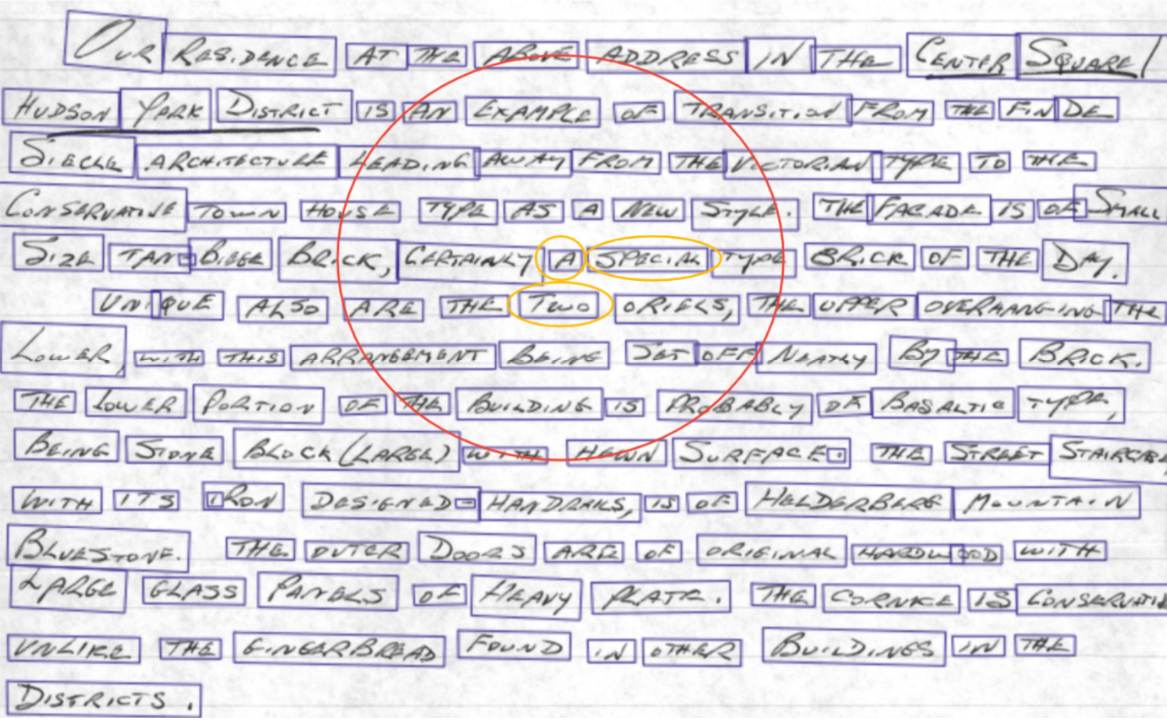}
    \caption{Consider the three words in orange circles. To predict any property that relates to the box that contains “A”, a linear-sized GCN can only use information in its neighborhood. For example, a GCN by itself, using information only in the red circle, can determine whether “A” is in the same line as “SPECIAL”, but cannot determine whether “A” is in the same paragraph as “Two”.
}
    \label{fig:gcn_nbhd}
\end{figure}

Consider a naive attempt that is to simply predict for each GCN edge, whether the two GCN boxes connected by the edge are from the same paragraph. Figure~\ref{fig:gcn_nbhd} shows that this would not work as we had expected --- Humans typically infer paragraph structures by tracing all the way back to the beginning of each line, no matter how long the line is; on the other hand, each edge in a linear-sized GCN can only receive information from a neighbourhood of the edge.

We overcome this challenge by making the GCN to predict only local relations and later using these local relations to perform global inference. See Section~\ref{sec:gcn_prediction} for a detailed discussion.

\subsection{$\beta$-skeleton Graph with 2-hop Connections}
\label{sec:gcn_edges}

\begin{figure}
    \centering
    \includegraphics[width=0.8\linewidth]{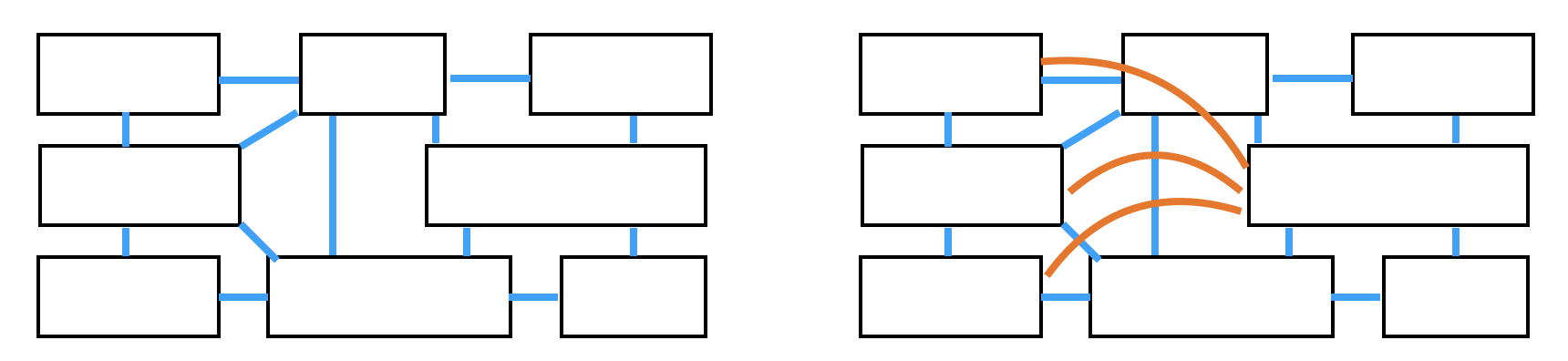}
    \caption{Left: $\beta$-skeleton graph on a set of word boxes \cite{post_ocr}. Right: Added 2-hop edges so that neighboring words within the same line are connected.}
    \label{fig:hop_edges}
\end{figure}

The $\beta$-skeleton graph used in \cite{post_ocr} is a good candidate as an efficient graph for layout tasks. However, in some cases with a combination of low line spacing and high word spacing, the graph constructed on word boxes cannot guarantee connectivity within all lines. As shown on the left of Figure~\ref{fig:hop_edges}, the middle line has an extra large space between the two words, and the $\beta$-skeleton edge in this space is a vertical cross-line connection rather than a horizontal edge we need for line-level clustering.

The solution we use is to add 2-hop connections to the graph. For each node in the $\beta$-skeleton graph, we check each of its neighbor's neighbor and add a hop edge if the two boxes fall in certain distance and angle constraints. This provides the necessary connectivity while still maintaining reasonable sparsity (shown in Figure~\ref{fig:hop_edges} on the right). Also see the graph on the right of Figure~\ref{fig:gcn_boxes_and_edges} for an example from synthetic documents.

\subsection{GCN Predictions}
\label{sec:gcn_prediction}

\begin{figure}
    \centering
    \includegraphics[width=\linewidth]{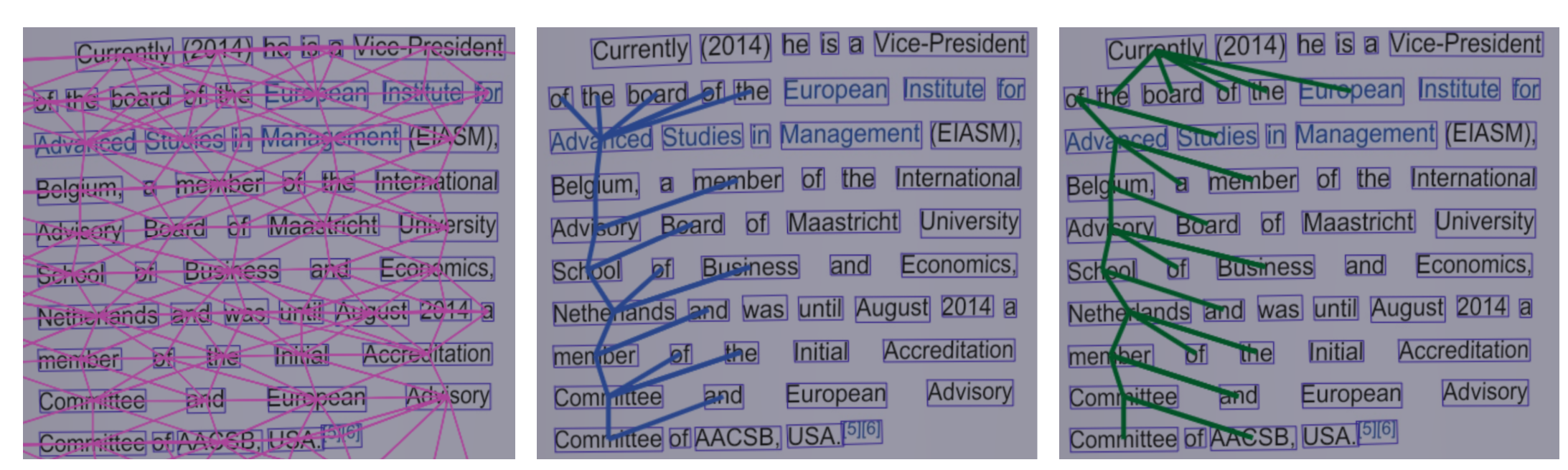}
    \caption{The left image shows the graph edges. \boxbelow~could miss the connection between the first two lines (middle image), but this could be complemented by \boxabove~(right image).}
    \label{fig:boxbelow_miss}
\end{figure}

\begin{figure}
    \centering
    \includegraphics[width=\linewidth]{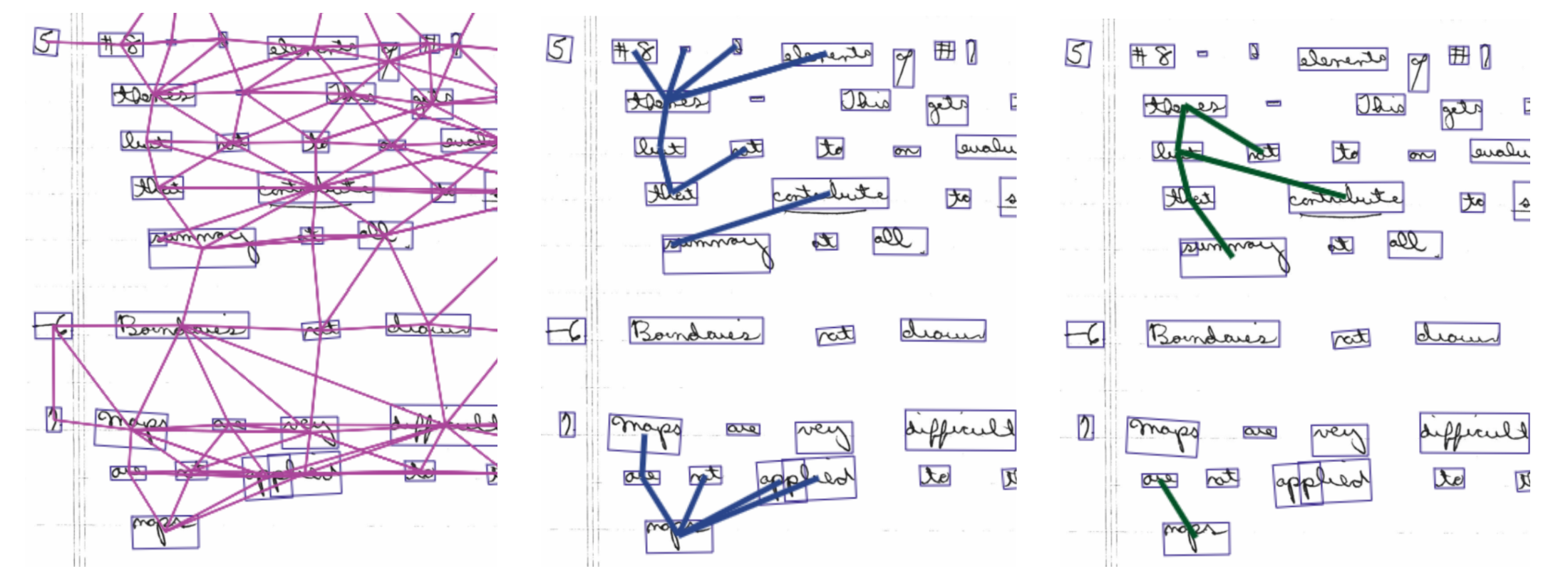}
    \caption{The left image shows the graph edges. \boxabove~could miss the connection between the first two lines of paragraphs (right image), but this could be complemented by \boxbelow~(middle image).
}
    \label{fig:boxabove_miss}
\end{figure}

Let $b_1$ and $b_2$ be two GCN boxes that are connected by a GCN edge. We use a GCN to make the following binary predictions.
\begin{itemize}
    \item 
    $ \boxleft(b_1, b_2)$: 1 if $b_1$ and $b_2$ are in the same line and $b_1$ is adjacent to $b_2$ to the left, 0 otherwise (Figure~\ref{fig:box_left}).
    \begin{figure}
        \centering
        \includegraphics[width=0.9\linewidth]{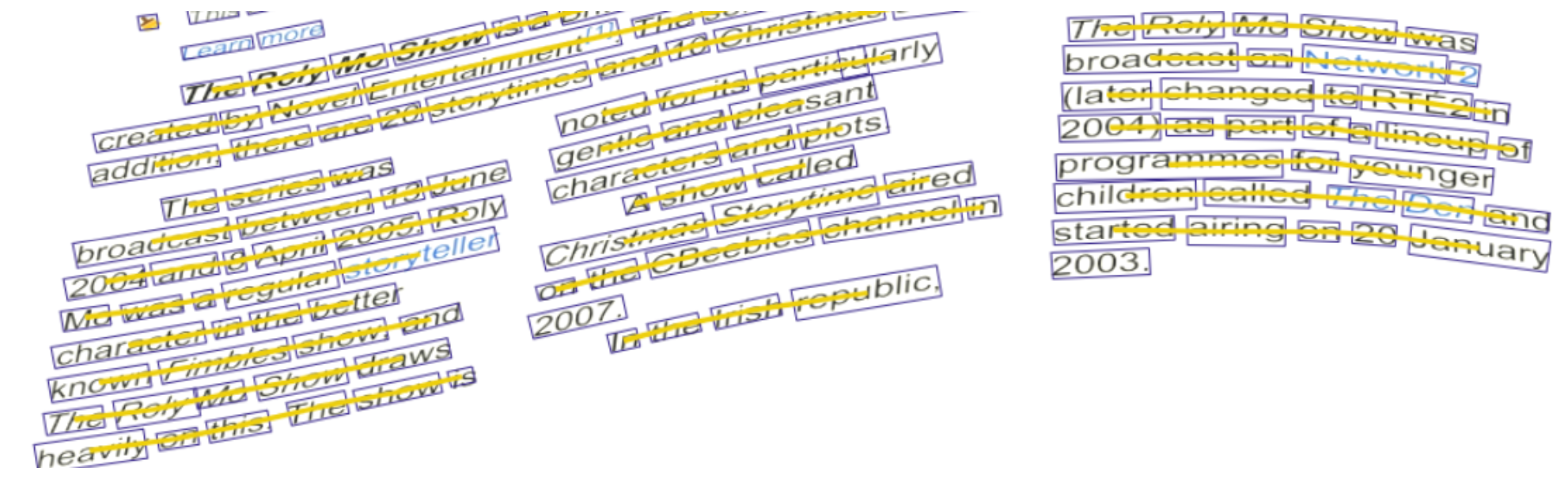}
        \caption{GCN boxes and \boxleft~labels. If there is a yellow edge connecting two GCN boxes $b_1$ on the left and $b_2$ on the right, it means $\boxleft(b_1, b_2) = 1$.}
        \label{fig:box_left}
    \end{figure}
    \item
    $ \boxabove(b_1, b_2)$:  (only makes valid prediction when $b_1$ is the first box in a line) 1 if $b_1$ and $b_2$ are in the same paragraph and $b_1$ is in a line that is directly above the line $b_2$ is in (Figure~\ref{fig:box_above}).
    \begin{figure}
        \centering
        \includegraphics[width=0.9\linewidth]{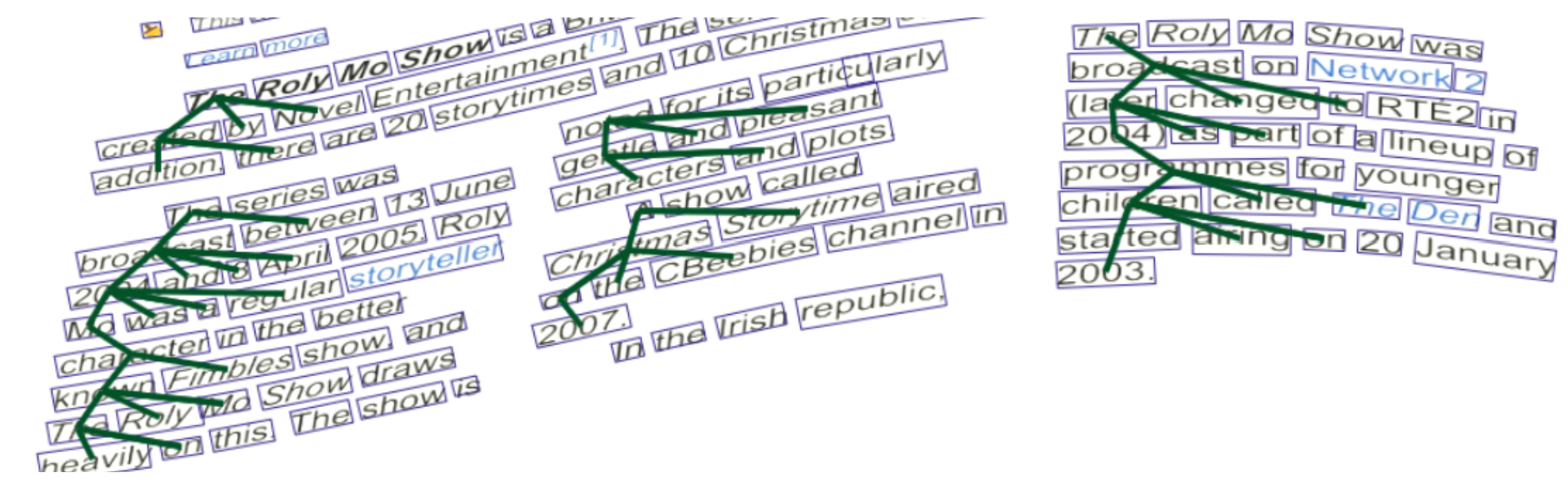}
        \caption{GCN boxes and \boxabove~labels. If there is a green edge connecting two GCN boxes $b_1$ on the top and $b_2$ on the bottom, it means $\boxabove(b_1, b_2) = 1$.}
        \label{fig:box_above}
    \end{figure}
    \item
    $ \boxbelow(b_1, b_2)$:  (only makes valid prediction when $b_1$ is the first box in a line) 1 if $b_1$ and $b_2$ are in the same paragraph and $b_1$ is in a line that is directly below the line $b_2$ is in (Figure~\ref{fig:box_below}).
    \begin{figure}
        \centering
        \includegraphics[width=0.9\linewidth]{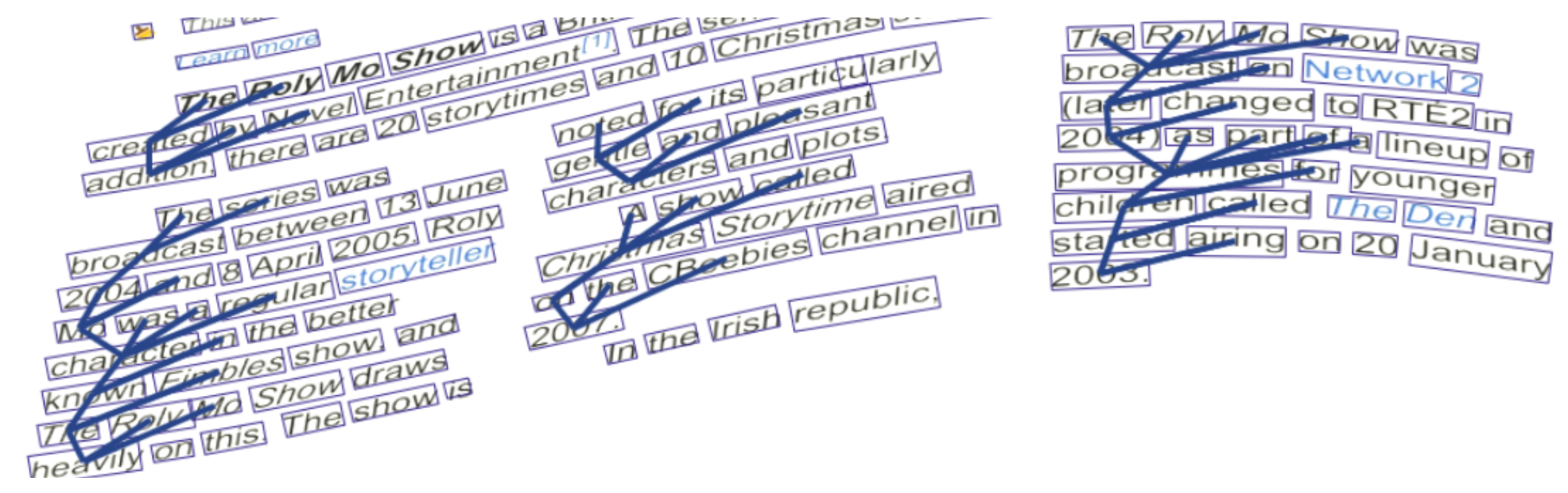}
        \caption{GCN boxes and \boxbelow~labels. If there is a blue edge connecting two GCN boxes $b_1$ on the bottom and $b_2$ on the top, it means $\boxbelow(b_1, b_2) = 1$.}
        \label{fig:box_below}
    \end{figure}
\end{itemize}
A curious reader may wonder whether it is necessary to have both \boxabove~and \boxbelow. When GCN edges has size linear in the size of GCN boxes, we cannot guarantee that the first GCN boxes in two adjacent lines are always connected by a GCN edge, especially when there is a large indentation. By utilizing \emph{both} \boxabove~and \boxbelow, we can be assured that the first GCN box in each line is connected to \emph{some} GCN box in any adjacent line. Figure~\ref{fig:boxbelow_miss} gives an example where \boxbelow~misses such a connection but \boxabove~does not; figure~\ref{fig:boxabove_miss} gives an example where \boxabove~misses such a connection but \boxbelow~does not.

\subsection{Forming Lines}
Initially all GCN boxes belong to a separate line. Iteratively, if $\boxleft(b1, b2) = 1$, then we merge $b_1$ and $b_2$ into the same line. A GCN box $b$ is identified as the first box of a line if there is no GCN box $b’$ such that $\boxleft(b’, b) = 1$.

\subsection{Forming Paragraphs}
Initially all lines belong to a separate paragraph. Iteratively, if $\boxabove(b1, b2) = 1$ or $\boxbelow(b1, b2) = 1$, then we merge $b_1$’s line and $b_2$’s line into the same paragraph.

\subsection{Overall System Pipeline}
\begin{figure}
    \centering
    \includegraphics[width=\linewidth]{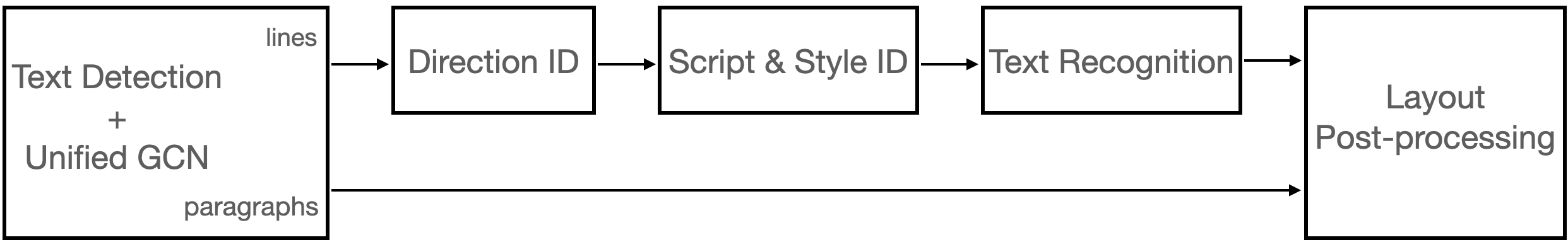}
    \caption{The overall OCR pipeline with the unified GCN model producing both lines and paragraphs at text detection.
}
    \label{fig:ocr_pipeline}
\end{figure}

The end-to-end pipeline of our OCR system with the proposed unified GCN model is shown in Figure \ref{fig:ocr_pipeline}. It is similar to the pipeline in \cite{handwritten}, with extra paragraph outputs directly produced in the text detector, so the layout post-processing can be greatly simplified, resulting in better system efficiency.

\section{Limitations}
\label{sec:limitations}

\subsection{Single-Line Paragraphs}
\label{sec:single_line}
\begin{figure}
    \centering
    \includegraphics[width=0.7\linewidth]{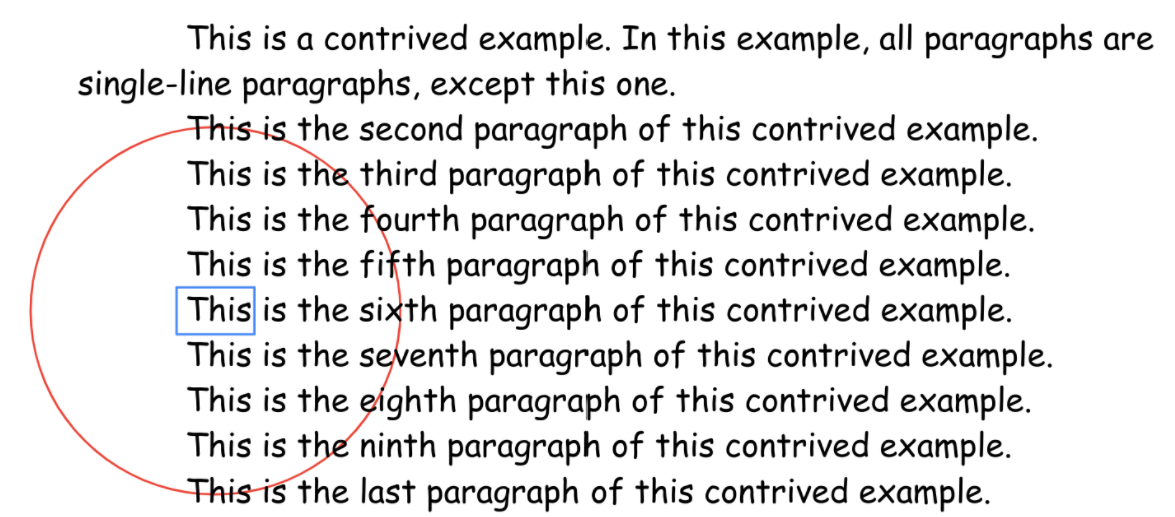}
    \caption{In rare cases, the limited perceptive field of a linear-sized GCN (red) could make it unable to determine whether a GCN box (blue) starts a paragraph.
}
    \label{fig:contrived}
\end{figure}

In rare circumstances, there are many consecutive equal-lengthed paragraphs and a linear-sized GCN would fail to identify the signal of a paragraph start from local information (Figure~\ref{fig:contrived}). This is rather a limitation of linear-sized GCNs and is beyond the scope of the current project.

\subsection{Document Rotations}

Since our approach is purely geometric, we focus on document images that are rotated no more than 45 degrees. GCN models are shown in \cite{post_ocr} to be much more robust than object detection models for rotated/distorted inputs.

However, when the rotation angle gets close to 90 degrees or even 180 degrees, the document orientation becomes uncertain, which will affect the \boxabove{} and \boxbelow{} predictions for paragraph-level clustering at the left side of text regions.

\begin{figure}
    \centering
    \includegraphics[width=0.6\linewidth]{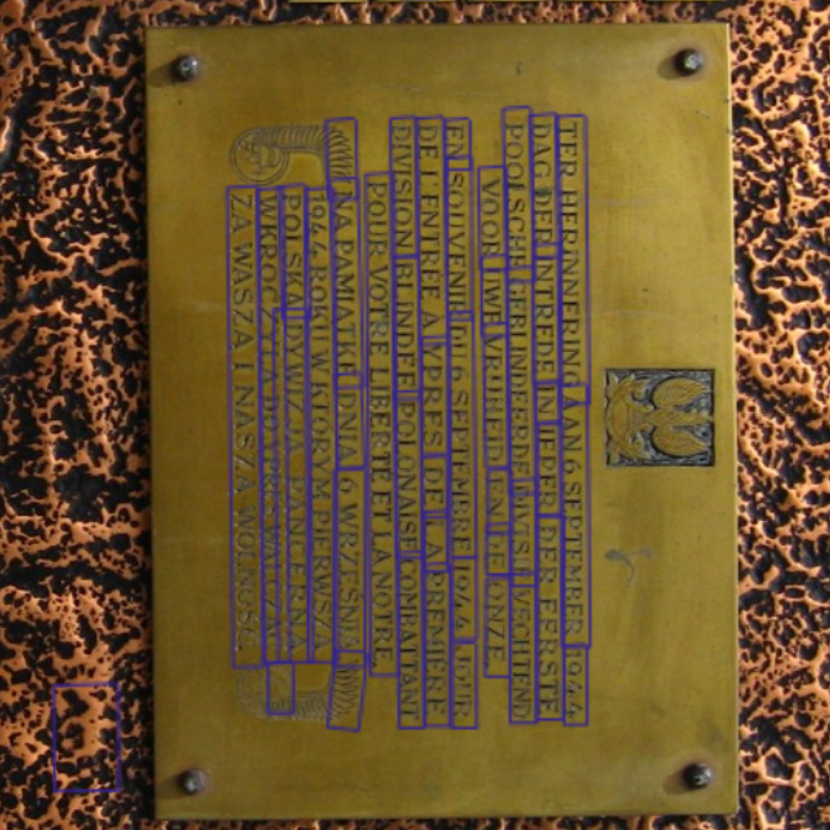}
    \caption{Image example rotated by 90 degrees, where it is hard to tell whether the texts are rotated clockwise or counter-clockwise purely from geometric information.}
    \label{fig:rotated_image}
\end{figure}

As shown in Figure~\ref{fig:rotated_image}, if a document image is rotated by 90 degrees, it may be impossible, even for a human, to recover the original orientation correctly using only geometrical information (bounding boxes). In a typical OCR engine, text directions can be obtained from a ``direction ID'' model~\cite{handwritten} which runs on detected text lines (see Figure \ref{fig:ocr_pipeline}). But here we don't have text lines yet, and will run into a ``chicken-and-egg'' dependency cycle to rely on line based text direction models. So if directions are needed, we may add a direction prediction to each character-level detection box, and take majority votes to decide the document orientation.

\section{Experiments}

We experiment with the proposed GCN model on both the open PubLayNet dataset \cite{publaynet} and our own annotated paragraph set from real-world images. These datasets can evaluate the clustering quality on output paragraphs, and on output lines as well since any mistake in line-level clustering will affect the parent paragraph.

The model input is from the OCR text detector behind the Google Cloud Vision API of DOCUMENT\_TEXT\_DETECTION~\footnote{https:$\sslash$cloud.google.com/vision/docs/fulltext-annotations} version 2021. Following the input layer are 5 steps of ``message passing'' in \cite{mpnn} as the backbone and 3 binary classification heads for the outputs described in section \ref{sec:gcn_prediction} -- similar to the ``line clustering'' model in \cite{post_ocr} but with 3 independent node-to-edge prediction heads.

In each message passing step, the node-to-edge message kernel has a hidden layer of size 64 and output layer of size 32, the edge-to-node aggregation is by average pooling, the node-to-next-step and the final prediction heads have the same (64, 32) sizes as the node-to-edge message kernel. The resulting GCN model is under 200KB in parameter size, which is negligible compared to a typical OCR engine or an image based layout model.

\subsection{PubLayNet Results}

We first train the unified GCN model using labels inferred from the PubLayNet training set, and evaluate the paragraph results on the PubLayNet validation set against the F1$_{var}$ metric introduced in \cite{post_ocr}. The result comparison with a few other methods are shown in Table~\ref{tab:publaynet}.

\begin{table}
\centering
\caption{Paragraph F1$_{var}$ score comparisons.}
\label{tab:publaynet}       
\begin{tabular}{lc}
\noalign{\smallskip}\noalign{\smallskip}
\hline \hline\noalign{\smallskip}
Model & F1$_{var}$ \\
\noalign{\smallskip}\hline\noalign{\smallskip}
Tesseract~\cite{tesseract} & 0.707 \\
Faster-RCNN-Q~\cite{post_ocr} & 0.945 \\
OCR + Heuristic~\cite{post_ocr} & 0.364 \\
OCR + 2-step GCNs~\cite{post_ocr} \hspace{10mm} & 0.959 \\
Unified GCN  & 0.912 \\
\noalign{\smallskip}\hline
\end{tabular}
\end{table}

\begin{figure}[H]
    \centering
    \includegraphics[width=\linewidth]{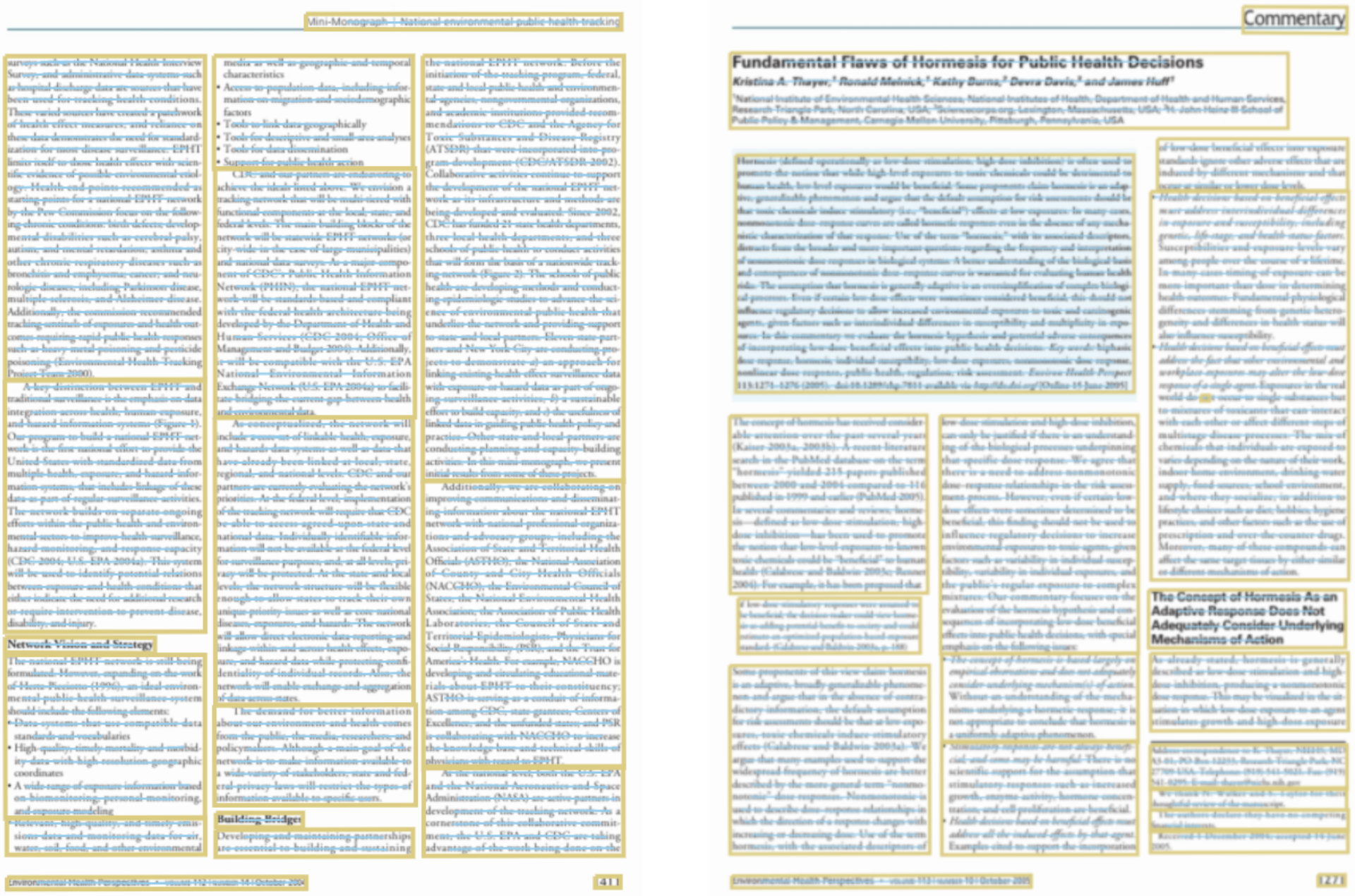}
    \caption{Evaluation examples from PubLayNet \cite{publaynet}. Paragraph-clustering results are mostly perfect except for list items (not annotated as paragraphs in ground truth) and the title section on top-right (not separating the title and author names due to lack of indentations). }
    \label{fig:publaynet}
\end{figure}

Compared to the results of 2-step GCNs, the only obvious loss pattern is caused by a situation discussed in section~\ref{sec:single_line}, which appears frequently in the title sections of this dataset, as shown in the top-right of Figure~\ref{fig:publaynet}. The paragraphs here are not indicated by indentation or vertical spacing, but by changes in font size and style which are not easily captured by early stage text detection. The 2-step approach can do better in this case because the line clustering model has more useful input features such as the total length of each line.

\subsection{Real-world Evaluation Results}

For evaluation on real-world images, we train the unified GCN model with both the augmented web synthetic data from \cite{post_ocr} and human annotated data. The metric is F1@IoU0.5 since there is no ground truth of paragraph line count to support the variable IoU threshold. Results are in Table~\ref{tab:real_world}.

\begin{table}
\centering
\caption{Paragraph F1-scores tested on the real-world test set.}
\label{tab:real_world}       
\begin{tabular}{llc}
\noalign{\smallskip}\noalign{\smallskip}
\hline \hline\noalign{\smallskip}
Model & \multicolumn{2}{r}{Training Data \hspace{20mm} F1@IoU0.5} \\
\noalign{\smallskip}\hline\noalign{\smallskip}
OCR + Heuristic & - &  0.602 \\
\noalign{\smallskip}
Faster-RCNN-Q \cite{post_ocr} & Annotated data & 0.607 \\
&  (pre-trained on PubLayNet) & \\
\noalign{\smallskip}
OCR + 2-step GCNs \cite{post_ocr} \hspace{5mm} & Augmented web synthetic + Annotated & 0.671 \\
\noalign{\smallskip}
Unified GCN & Augmented web synthetic + Annotated \hspace{5mm} & 0.659 \\
\noalign{\smallskip}\hline
\end{tabular}
\end{table}

\begin{figure}[H]
    \centering
    \includegraphics[width=\linewidth]{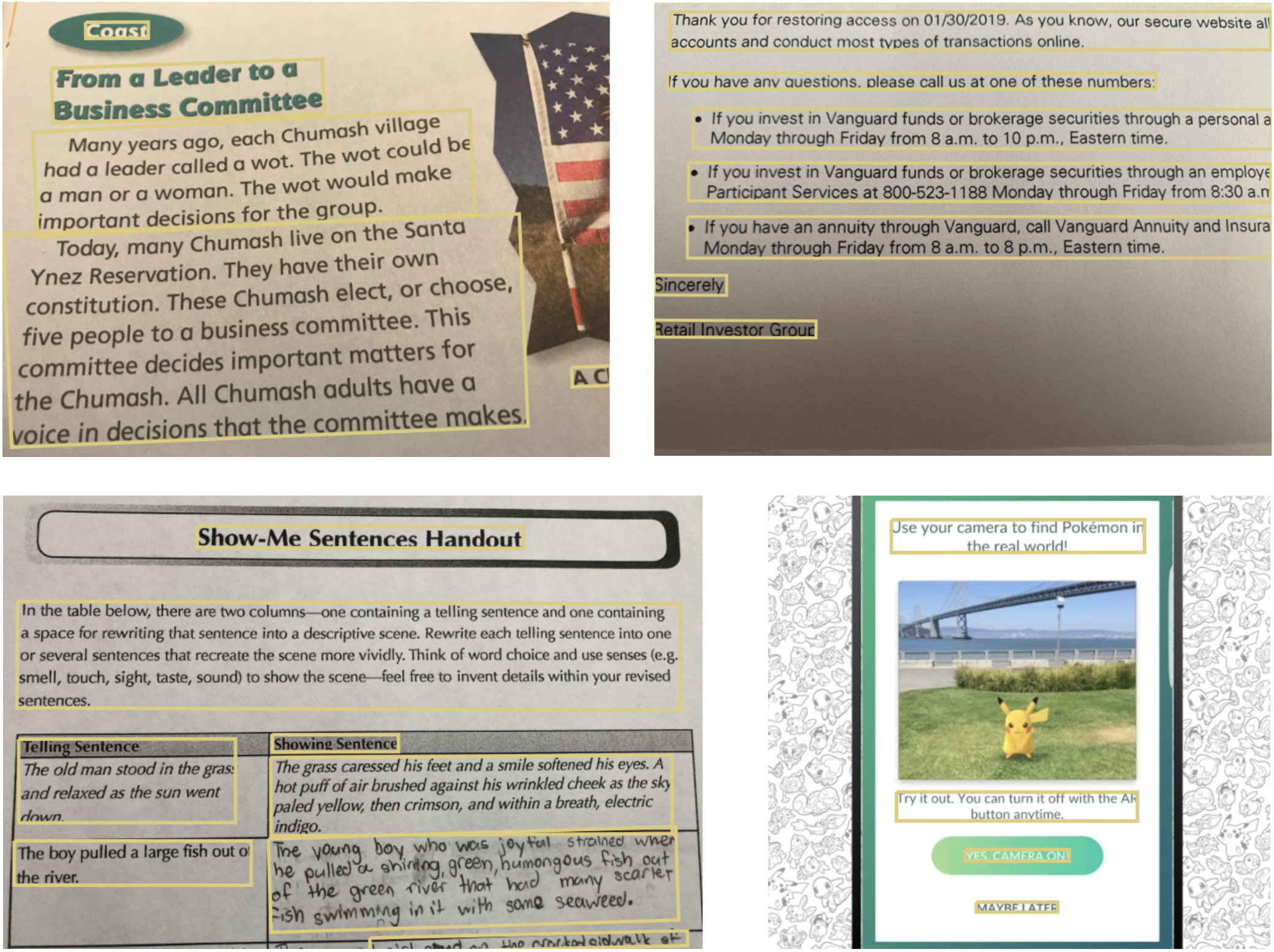}
    \caption{Success examples among real-world images.}
    \label{fig:realworld_success}
\end{figure}

\begin{figure}[H]
    \centering
    \includegraphics[width=\linewidth]{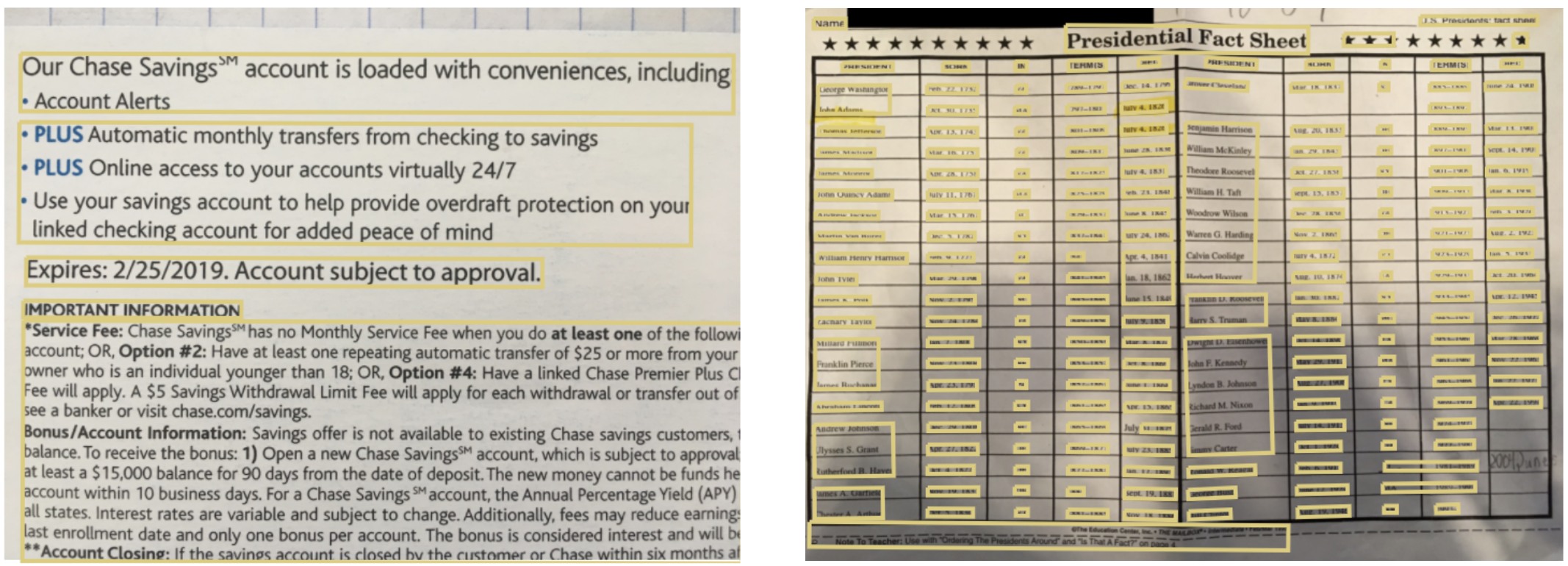}
    \caption{Failure examples among real-world images. Left: paragraphs without indentations or extra vertical spacing may not be detected. Right: tables will need extra handling beyond the scope of this work.}
    \label{fig:realworld_fail}
\end{figure}

Figure~\ref{fig:realworld_success} and \ref{fig:realworld_fail} show success and failure examples from this set. The loss pattern from section~\ref{sec:single_line} is still here in the left image of Figure~\ref{fig:realworld_fail}. But such cases are quite rare in the real world as told by the smaller score difference than in Table~\ref{tab:publaynet}. The overall quality of the unified GCN output is very close to that from the 2-step approach.

\section{Conclusions and Future Work}

We demonstrate that two OCR related layout tasks can be performed by a single unified graph convolutional network model. Compared to a multi-step pipeline, this unified model is more efficient and can produce paragraphs at virtually the same quality in real world images.

\begin{figure}[H]
    \centering
    \includegraphics[width=0.6\linewidth]{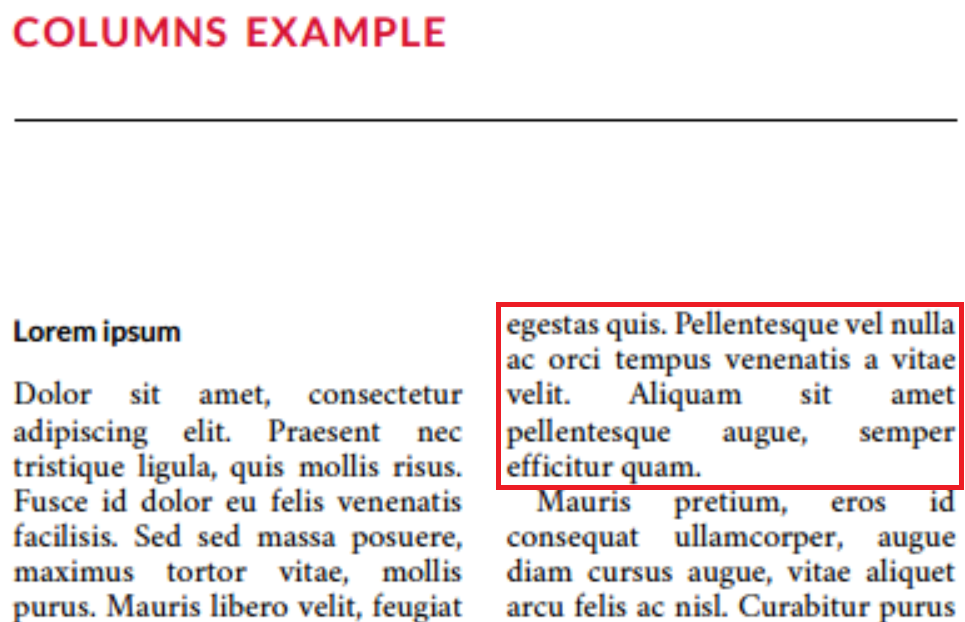}
    \caption{The region contained in the red rectangle is physically a paragraph, but needs to be combined into the one in the left text column to form a complete semantic paragraph.}
    \label{fig:false_start}
\end{figure}

We believe the number of clustering levels is not limited to two, since document layout tasks are extremely diverse in nature. Paragraphs can further be clustered into text columns or sections, which may belong to even higher level blocks. Figure~\ref{fig:false_start} shows a physical paragraph, or part of a semantic paragraph which spans across multiple text columns. Our current approach stops at physical paragraphs. But if we can obtain text columns as layout entities, semantic paragraphs will be available to provide better document structures for downstream applications. Future work will further study unified approaches for such layout problems.

\bigskip
\noindent\textbf{Acknowledgements.} The authors would like to thank Reeve Ingle and Ashok C. Popat for their helpful reviews and feedback.

\end{document}